\documentclass[sigplan,screen]{acmart}

\renewcommand\footnotetextcopyrightpermission[1]{}
\usepackage{amsfonts}
\usepackage{algorithmic}
\usepackage{graphicx}
\usepackage{textcomp}
\usepackage{xcolor}
\usepackage{colortbl}
\usepackage{ulem}
\usepackage{url}
\usepackage{subcaption} 
\usepackage{multirow}
\usepackage{balance}
\usepackage{makecell}
\usepackage{enumitem}
\usepackage[usenames,dvipsnames]{xcolor}

\newcommand\greencheck{{\color{OliveGreen}\checkmark}}
\newcommand\redx{{\color{red}$\times$}}

\newcommand{\tool}{\texttt{DiTOX}}

\title{DiTOX: Fault Detection and Localization in the \mbox{ONNX Optimizer}}
\titlenote{The research has been funded by Huawei Edinburgh Joint Lab Project "RobustCheck: Testing Robustness of Compiler Optimisations and Deep Learning Frameworks".}

\author{Nikolaos Louloudakis}
\affiliation{
  \institution{University of Edinburgh}
  \department{School of Informatics}
  \city{Edinburgh}
  \country{United Kingdom}
}
\email{n.louloudakis@ed.ac.uk}

\author{Ajitha Rajan}
\affiliation{
  \institution{University of Edinburgh}
  \department{School of Informatics}
  \city{Edinburgh}
  \country{United Kingdom}
}
\email{arajan@ed.ac.uk}

\acmConference[CC '26]{Proceedings of the 35th ACM SIGPLAN International Conference on Compiler Construction}{January 31 -- February 1, 2026}{Sydney, NSW, Australia}
\acmBooktitle{Proceedings of the 35th ACM SIGPLAN International Conference on Compiler Construction (CC '26), January 31 -- February 1, 2026, Sydney, NSW, Australia}
\received{2025-11-03}
\received[accepted]{2025-12-10}



\begin{document}


\begin{abstract}

The ONNX Optimizer, part of the official ONNX repository and widely adopted for graph-level model optimizations, is used by default to optimize ONNX models. Despite its popularity, its ability to preserve model correctness has not been systematically evaluated. We present \tool, an automated framework for comprehensively assessing the correctness of the ONNX Optimizer using differential testing, fault localization, and evaluation techniques that generalize to other compiler optimizers. \tool\ applies optimization passes to a corpus of ONNX models, executes both original and optimized versions on user-defined inputs, and detects discrepancies in behavior or optimizer failures. When divergences are observed, \tool\ isolates the responsible optimization pass through iterative, fine-grained analysis. We evaluated \tool\ on 130 models from the ONNX Model Hub spanning vision and language tasks. We found that 9.2\% of model instances crashed the optimizer or produced invalid models under default settings. Moreover, output discrepancies occurred in 30\% of classification models and 16.6\% of object detection and segmentation models, while text-based models were largely robust. Overall, \tool\ uncovered 15 issues—14 previously unknown—affecting 9 of the 47 optimization passes as well as the optimizer infrastructure. All issues were reported to the ONNX Optimizer developers. Our results demonstrate that \tool\ provides a simple and effective approach for validating AI model optimizers and is readily extensible beyond ONNX.
\end{abstract}

\maketitle


\section{Introduction}
Deep Learning (DL), primarily implemented through Deep Neural Networks (DNNs), is applied across a wide range of domains, including autonomous vehicles and medical imaging. As these systems become more prevalent and are assigned increasingly complex tasks, the demand for their computational efficiency intensifies. To address this, newer and more powerful hardware devices and architectures are being developed, and effectively utilizing their performance depends on compilers and the optimizations they apply to deep learning models~\cite{speedisallyouneed, CompilerGym}. Widely-used AI compilers and frameworks, such as Apache TVM~\cite{tvm} and MLIR~\cite{lattner2020mlircompilerinfrastructureend}, incorporate their own native sets of optimization passes for efficient use of hardware. These include standard techniques such as operator fusion and constant folding, as well as more aggressive approaches like reduced-precision arithmetic, including the use of fast-math optimizations, which may compromise numerical accuracy. Models can also be represented using industry-standard formats, such as ONNX~\cite{onnxsite}. By applying transformations to models in these formats, it is possible to reduce model size and improve performance. For ONNX in particular, the ONNX Optimizer~\cite{onnxoptimizer} serves as the default tool for performing graph-level optimizations. As part of the official ONNX toolkit, it focuses on simplifying model structure while aiming to preserve functional correctness, particularly in terms of accuracy. However, the degree to which these optimizations maintain model accuracy remains an open question.

We investigate the extent to which ONNX Optimizer transformations affect the functional correctness of DNN models. To this end, we propose the \textbf{Di}fferential \textbf{T}esting of the \textbf{O}NN\textbf{X} Optimizer (\tool)\footnote{The source code of \tool, along with usage instructions, is available at \emph{\url{https://github.com/luludak/DiTOX}}.}~\cite{DiTOX}, a suite that detects and localizes faults in the ONNX Optimizer.
\vspace{-7pt}
\paragraph{Problem Being Solved}
\tool\ solves the problem of detection and localization of faults associated with the ONNX Optimizer, primarily in the direction of output label inconsistencies but also compiler crashes and miscompilations.
In particular, \tool\ employs a per-pass fault localization strategy combined with differential testing between original and optimized models of various types to identify and localize issues arising from specific passes in the ONNX Optimizer.
\vspace{-7pt}
\paragraph{Case Studies} To evaluate the effectiveness of \tool, we collected 130 models as case studies, performing different tasks, such as classification, object detection, text comprehension, and text generation. All models were obtained from the official ONNX Model Hub~\cite{onnxmodelhub}. Each model was optimized using the ONNX Optimizer and assessed on standard benchmark datasets to compare accuracy before and after optimization. When discrepancies were detected, we applied pass-by-pass optimization to isolate the specific transformations responsible for the faults.

We focus on the ONNX Optimizer due to its impact and popularity (over 770 GitHub stars) as a model optimization tool. In addition, it is integrated into the main ONNX repository and serves as a foundation for other widely adopted utilities, such as the ONNX Simplifier~\cite{onnxsimplifier}, which has received over 4,000 GitHub stars.
However, existing ONNX Optimizer tests primarily focus on toy models rather than real-world models, primarily used for unit testing purposes. With \tool, we aim to enhance the effectiveness of fault localization and detection in the optimizer.
However approach is not limited to ONNX; it is generalizable to other AI compilers and optimization frameworks.

By conducting experiments on our case studies, we observed output discrepancies between the original and optimized models in $34$ of $130$ cases ($26.1$\%). We also encountered issues with $12$ model instances ($9.2$\%) using the primary optimizer passes (\texttt{fuse} and \texttt{eliminate}) where the optimizer crashed or produced invalid models. Through analysis, we identified 15 distinct bugs, only one of which had been partially reported before (without a full explanation or reproduction steps). For this case, we provided additional insights. All bugs were reported to the official ONNX Optimizer repository with detailed descriptions for further resolution. \tool\ effectively detected faults due to incorrect value references, input/output mismatches, and graph-structure inconsistencies introduced by the ONNX optimizer. These manifested as (1) optimizer crashes, (2) malformed models, and (3) valid models with output label inconsistencies compared to their unoptimized counterparts.

\vspace{-10pt}
\section{Background}
\subsection{Deep Learning Models}
Deep Neural Networks (DNNs) are widely used for prediction tasks across domains. We focus on three common categories:
\subsubsection{Image Classification}
Image classification, one of the earliest DNN applications, typically uses Convolutional Neural Networks (CNNs) to extract image features via learned convolution filters. These models output a ranked list of top-K predictions, each paired with a confidence score, typically normalized using a Softmax function~\cite{softmax}.
\subsubsection{Object Detection and Semantic Segmentation}
Object detection extends classification by identifying object locations using bounding boxes. Semantic segmentation further refines this by labeling each pixel with a class (e.g., road, pedestrian), offering dense scene understanding. Both rely heavily on convolutional operations and are critical in areas like autonomous driving.
\subsubsection{Text Comprehension and Generation}
These models handle tasks such as summarization, recognition, and sentiment analysis. Text generation models, especially transformers~\cite{attention}, use contextual representations to iteratively predict the next token using a sliding window approach. This generative process continues until a stopping condition is met.

\vspace{-5pt}
\subsection{Graph Optimizations}
For deep neural networks (DNNs) to perform effectively on complex tasks, they must be optimized to operate more efficiently while maintaining their performance. A common approach to achieving this is optimizing the model graph to improve computational efficiency. This may involve techniques such as node pruning, node fusion, and constant folding, among others. Similar to conventional software, these optimizations are applied to DNNs, with AI compilers~\cite{tvm, lattner2020mlircompilerinfrastructureend} incorporating such techniques. Additionally, tools like the ONNX Optimizer, which we examine in this work, are specifically designed for this purpose.

\subsection{The ONNX Optimizer}
\label{sub:onnx-optimizer}
The ONNX Optimizer is a graph-level optimization tool designed to improve the performance and portability of ONNX models by applying a series of transformation passes. It consists of a set of $47$ passes\footnote{\label{passesfootnote}The complete list of passes, each accompanied by a brief description, can be found at  \emph{\url{https://github.com/luludak/DiTOX/blob/master/passes.txt}}.} grouped into three main types: fusions, eliminations, and graph rewrites, with the purpose of improving model efficiency. Fusion passes merge operations (e.g., fuse\_bn\_into\_conv, which merges \texttt{BatchNorm} and \texttt{Conv} nodes) to reduce overhead. Elimination passes remove redundant nodes, such as \texttt{eliminate\_deadend} (removing nodes with unused outputs) and \texttt{eliminate\_nop\_transpose} (removing transpose nodes that do nothing). Graph rewrite passes modify the graph structure to ensure compatibility; for example, \texttt{rewrite\_input\_dtype} adjusts input data types. In this category, there are also secondary passes designed primarily for analysis and debugging, such as \texttt{split\_init} and \texttt{split\_predict}, which isolate the initializer and inference subgraphs, respectively. By default, the optimizer applies only fusion and elimination passes; however, it also supports applying any subset of its $47$ transformation passes, based on its user-defined configuration.

\subsection{Model Output Comparison Methods}

Based on the model and output type, \tool\ uses a set of well-known and established methods in order to compare the outputs of the optimized model versions with their original counterparts. In particular:

\begin{description}[leftmargin=0pt]
    \item[Classification models:] 
    We compare the top-K label predictions between original and optimized models using Kendall's Tau Rank Correlation~\cite{kendall}, which evaluates the similarity between ranked label lists, with scores ranging from -1 (inverse correlation) to 1 (perfect correlation).
    \vspace{5pt}
    \item[Object Detection and Semantic Segmentation models:]{\hfill
    Each prediction includes (1) class probabilities, (2) bounding box coordinates, and (3) a confidence score. For class probabilities, we apply a softmax and use Kendall’s Tau to compare rank similarity. For bounding boxes and confidence scores, we use standard metrics: (I) IoU for bounding box overlap, (II) F1 Score for overall detection balance, (III) Average Precision (AP) for precision across thresholds, (IV) Average Recall (AR) for recall across thresholds, (V) mAP for aggregate detection and localization accuracy. These metrics are computed at IoU thresholds of $0.5$, $0.75$, and $0.9$, with the original model as the reference.}
    \vspace{5pt}
    \item[Text generation models:] 
    We convert token outputs to plain text and use the BLEU Score~\cite{BLEU} to compare the optimized model's output against the original, assessing n-gram overlap and overall similarity.
    \vspace{5pt}
    \item[Binary sentiment analysis models:] We measure the percentage of prediction label differences between original and optimized models to evaluate the impact of optimization on classification consistency.
\end{description}

\vspace{-10pt}
\section{System Architecture \& Methodology}
\label{system-arch}

\begin{figure*}[!htp]
\centering
\includegraphics[width=\textwidth]{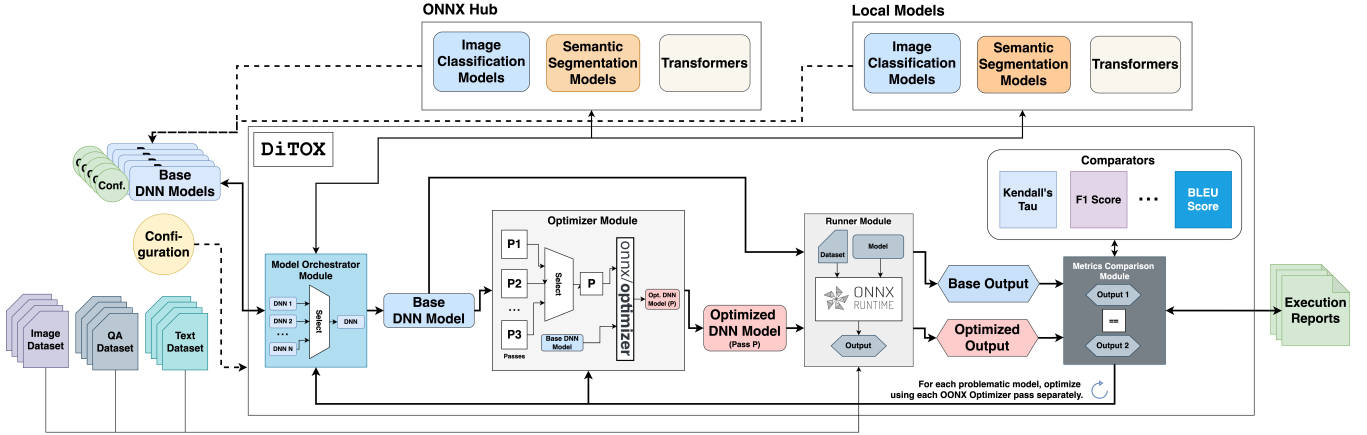}
 \vspace{-5pt}
 \caption{Architecture of \tool\ with the following primary modules: (1) Model Orchestrator, (2) Optimizer Module, (3) Runner Module, and (4) Metrics Comparison Module.}
 \label{fig:DiTOX}
 \vspace{-10pt}
\end{figure*}

The architecture of \tool\ and its modules is shown in Figure~\ref{fig:DiTOX}. Its workflow is straightforward: load models, optimize them, execute both original and optimized versions across the dataset, and compare the results. If output discrepancies are detected, \tool\ iteratively re-optimizes each model using individual ONNX Optimizer passes to localize the pass causing the difference. All results are stored in report files for analysis and fault localization.

\tool\ consists of four primary modules: (1) Model Orchestrator, (2) Optimizer Module, (3) Runner Module, and (4) Metrics Comparison Module. It loads model configuration settings either (1) from a local configuration file or (2) by fetching them from the ONNX repository from which the models are sourced, both managed by the Model Orchestrator Module. The modules in Figure~\ref{fig:DiTOX} are detailed in the following sections.

\tool\ currently isolates bugs at the level of individual passes, relying on compiler error messages to identify the cause. For more effective fault localization, future enhancements can explore data flow analysis and delta debugging within the identified pass to identify the root cause.

\subsection{Model Orchestrator Module}
The Model Orchestrator Module is responsible for all tasks related to fetching, loading (including model configuration) and preparing models for optimization and execution. The module supports two loading options for DNN models: (1) local loading of ONNX models, and (2) loading models from an online repository. At the moment, \tool\ supports the official ONNX Model Hub~\cite{onnxmodelhub}, however, as it utilizes the official Hub API, it can trivially support third party ONNX model repositories. In addition, the module supports fetching and loading of multiple models simultaneously, while it also performs model caching when fetching from a model hub, as well as checksum verification.
Once each model and its respective configuration are properly loaded, they are propagated to the Optimizer Module and the Runner Module.

\vspace{-10pt}
\subsection{Optimizer Module}
The Optimizer module is responsible for the generation of optimized models, using the ONNX Optimizer at its heart. It supports two settings: (1) full optimization, applying all the ONNX Optimizer passes to the model under test, and (2) partial optimization, where only a select set of passes is applied to the model, applied in a user-defined order specified in the system configuration. It is worth noting that in (1), by optimizer design, only the \texttt{fuse} and \texttt{eliminate} passes are applied as part of the default optimization bundle in the ONNX Optimizer, leaving out a number of secondary passes (e.g., `rename\_input\_output`). \tool\ supports the utilization of these passes for testing, in order to explore all optimizer operations. This feature is essential throughout the \tool\ pipeline: when faulty behavior is discovered during model evaluation, the models exhibiting such faults must be examined in depth. To identify the pass responsible for problematic behavior, \tool\ first compares fully optimized models with their originals throughout its pipeline, and upon detecting a crash or accuracy issue, isolates the problematic cases by generating per-pass optimizations and comparing them again with the originals in an iterative process, until the pass responsible for the issue is identified.

Once a model is optimized with either setting, it then gets propagated to the Runner model for inference.

\vspace{-10pt}
\subsection{Runner Module}
The Runner model does precisely what its name suggests: executes the models under test. In particular, it receives both the original and the optimized model under test, as well as a dataset that will be used as input for both model versions. It then performs model inference, using ONNX Runtime~\cite{onnxruntime}, combined with the necessary model preprocessing, while utilizing the model configuration loaded in the Model Orchestrator component.
\tool\ supports different types of datasets, based on the type of models being examined - from images to IMDB reviews. It can load the inputs from local files, but also fetch the dataset from well-known packages, such as the \texttt{datasets}~\cite{datasets} package.

To enable the use of large-scale datasets and extract results from models that are larger and more computationally demanding—especially when operating on low-end devices—the Runner Module supports execution across a dataset in parts. In particular, a run can be divided into $N$ chunks, where each chunk consists of a fraction,  $1/N$th, of the dataset. Any remaining data not evenly divisible is also included in the final chunk of the run.
This process applies to both the original and the optimized model versions, with the results extracted and propagated to the Metrics Comparison Module for subsequent evaluation and reporting. By following this approach, we were able to conduct our survey on two consumer-grade devices without relying on a compute cluster to obtain results, as described in Section~\ref{s:experiment-setup}.

\subsection{Metrics Comparison Module}
\label{sub:comparison}
Once the run completes per-chunk for both the original and the optimized model, the results are propagated to the Metrics Comparison Module. This module performs comparison against the results, by selecting the appropriate comparator based on the model type. For classification types, \tool\ supports first-label comparison, but also calculates the Kendall Rank Correlation Coefficient across the $top-K$ predictions - commonly known as Kendall's Tau~\cite{kendall}. For Object Detection models, it utilizes the same methods of classification for labels and ranks, but also applies Intersection over Union (IoU~\cite{IoUBoundingBox}), Precision, Recall, and F1 Score~\cite{F1Score} given an IoU threshold. For text generation models (e.g., GPT-2), it utilizes BLEU~\cite{BLEU} score in order to compare the text generated out both the original and the optimized model, as well as direct tensor comparison for cases of question answering (Q\&A). In addition \tool\ is build in a manner that new comparators can be added, in cases that a new category of models is tested.
\vspace{-10pt}
\subsection{Fault Localization}
Once a method is selected, the Metrics Comparison Module calculates the metrics based on model type and generates a report with all results, in JSON format.
If a difference is detected across the original and the optimized model, then \tool\ retrofits the reported data to the Optimizer Module. Then, the Optimizer Module generates model optimizations per-pass, reporting how each pass behaves. \tool\ applies a fault localization methodology for three cases: when the optimizer crashes, when it generates an optimized model in an invalid form, and when the optimization succeeds but produces results inconsistent with the original model. In addition, \tool\ registers any warnings that can indicate the presence of a potential performance fault (e.g., redundant initializers in the model). Upon detecting an issue or a problematic behavior, \tool\ attempts to identify the responsible pass by generating per-pass optimizations, executing inference on the resulting models, and comparing their outputs with those of the original model. This process is applied iteratively for each of the passes supported by the optimizer, until the pass responsible for the detected issue is identified.


\vspace{-10pt}
\section{Related Work}
In this work, we compare our work with contributions related to AI compilers, focusing on the following key aspects: (1) the ability to localize faults associated with graph optimization passes, (2) the detection of compiler bugs affecting The types of bugs considered are the ones resulting in (a) model accuracy (inference results), (b) compiler crashes, (c) successful compilations, which, however, produce malformed or incorrect models that fail to run (e.g., models with incorrect input nodes or parameters), and (d) issues impacting execution times, (3) the capability to localize faults in the ONNX domain, particularly within ONNX Optimizer, and (4) the detection of faults in graph optimization passes. Table~\ref{t:related-work} summarizes the features supported by each contribution. For each tool, Table~\ref{t:related-work} shows: (1) the underlying technique, (2) support for bug detection in ONNX, (3) the ability to localize faults (\textit{Flt Loc.}), (4) the types of bugs detected and the corresponding graph optimization passes involved (\textit{Gr. Opt. Passes}), (5) detection of label accuracy faults (\textit{Acc.}), (6) detection of crash faults (\textit{Cr.}), (7) detection of malformed models produced despite successful compilation (\textit{Malf.}), (8) detection of bugs affecting execution times (\textit{Ex. T.}), and (9) the compilers supported/evaluated.

\pagenumbering{gobble}
\setlength{\tabcolsep}{1.5pt}
\begin{table*}[!htp]
     \small
    \centering
    \caption{Comparison of features between \tool\ and related work. Each technique is compared across the following parameters: testing of ONNX (\textit{ONNX?}), ability for fault localization (\textit{Flt. Loc}.), bug types, graph optimization passes (\textit{Gr. Opt. Passes}), detection of accuracy changes (\textit{Acc.}), crashes (\textit{Cr.}), generation of malformed models due to miscompilation (\textit{Malf.}), and issues affecting model execution times(\textit{T.}) bugs, as well as the compilers explored on each contribution.\\ $*:$ Limited support of feature, $\dagger:$ Does not consider ONNX Optimizer or compile-time optimizations associated with ONNX, $\ddag:$ Uses ONNX as an intermediate representation, but does not perform fault detection associated to it.}
    \begin{tabular}
    {|l|l|l|l|l|l|l|l|l|l|l|l|}
    \hline
        \textbf{Tool} & \textbf{Technique} & \textbf{\makecell[l]{ON\\NX?}} & \textbf{\makecell[l]{Flt.\\ Loc.}} & \textbf{Bug Types} & \textbf{\makecell[l]{Gr. Opt.\\ Passes}} & \textbf{Acc.} & \textbf{Cr.} & \textbf{Malf.} & \textbf{T.} & \textbf{Compilers} \\ \hline
        \textbf{\tool} & \makecell[l]{Per-pass fault\\ localization, \\differential \\testing} & ~\greencheck & ~\greencheck & \makecell[l]{Logic, API versioning,\\ graph input/parameters/\\transformations }& \makecell[l]{Fuse/\\eliminate/\\rewrite} & ~\greencheck & ~\greencheck & ~\greencheck & ~\greencheck$^{*}$ & \makecell[l]{ONNX  Optimizer~\cite{onnxoptimizer}} \\ \hline
        NNSmith~\cite{NNSmith} & \makecell[l]{Grammar-\\based fuzzing} & ~\greencheck$^{\dagger, \ddag}$ & ~\redx & \makecell[l]{Conversions, semantics,\\ transformations,\\ graph rewrites} & \makecell[l]{Conv2D\\ inputs\\ rewrite} & ~\greencheck$^{*}$& ~\greencheck & ~\redx & ~\redx & \makecell[l]{TVM~\cite{tvm},\\ PyTorch JIT~\cite{pytorch},\\ TensorRT~\cite{tensorrt},\\ PT2~\cite{pytorch2},  ONNX\\ Runtime~\cite{onnxruntime},\\ TF XLA~\cite{tfxla},\\ TFLite~\cite{tensorflow}} \\ \hline
        MLIRSmith~\cite{MLIRSmith} & \makecell[l]{Grammar-\\based fuzzing} & ~\redx$^{\dagger}$ & ~\redx & \makecell[l]{Conversions (dialects),\\ generic transformations,\\ logic (dialects)}  & ~\redx & ~\redx & ~\greencheck & ~\redx & ~\redx & \makecell[l]{MLIR~\cite{MLIR},\\ LLVM~\cite{LLVM}} \\ \hline
        MLIRod~\cite{SuoMLIR} & \makecell[l]{Operator\\ dependency/\\fuzzing} & ~\redx & ~\redx & \makecell[l]{Conversions (dialects),\\ transformations} & ~\redx & ~\redx & ~\greencheck & ~\redx & ~\redx & MLIR~\cite{MLIR} \\ \hline
        \makecell[l]{SYNTH-\\FUZZ~\cite{synthfuzz}} & \makecell[l]{Grammar-\\based fuzzing}  & ~\redx & ~\redx & Generic, input-based & ~\redx & ~\redx & ~\greencheck & ~\redx & ~\redx & MLIR~\cite{MLIR} \\ \hline
        HirGen~\cite{HirGen} & \makecell[l]{Graph IR\\ fuzzing} & ~\redx$^{\dagger}$ & ~\redx & \makecell[l]{Numeric computations,\\ exceptions, types, memory} & ~\redx & ~\redx & ~\greencheck & ~\redx & ~\redx & TVM~\cite{tvm} \\ \hline
        TVMFuzz~\cite{TVMFuzz} & \makecell[l]{Low-level\\ IR fuzzing} & ~\redx & ~\redx & \makecell[l]{Numeric computations,\\ graph rewrites} & \makecell[l]{Constant\\ folding} & ~\redx & ~\greencheck & ~\greencheck & ~\redx & TVM~\cite{tvm} \\ \hline
        Tzer~\cite{TZER} & \makecell[l]{Coverage-\\guided fuzzing \\(Low-level IR)} & ~\redx & ~\redx & \makecell[l]{API inconsistency/misuse,\\ memory, types} & ~\redx & ~\greencheck$^{*}$ & ~\greencheck & ~\redx & ~\greencheck$^{*}$ & TVM~\cite{tvm} \\ \hline
        GenCoG~\cite{GenCoG} & \makecell[l]{Constraint \\solving,\\ High-level\\ IR Fuzzing} & ~\redx & ~\redx & \makecell[l]{Memory, tensor shape\\/data, operator attributes,\\ logic} & \makecell[l]{Operator\\ fusion} & ~\greencheck$^{*}$ & ~\greencheck & ~\redx & ~\redx & TVM~\cite{tvm} \\ \hline
        DeepDiffer~\cite{DeepDiffer} & \makecell[l]{Low-level IR/\\coverage-\\guided fuzzing,\\ differential\\ testing} & ~\redx & ~\redx & \makecell[l]{Type, memory, API misuse,\\ numeric computations,\\ exception handling} & ~\redx & ~\redx & ~\greencheck & ~\redx & ~\greencheck$^{*}$ & TVM~\cite{tvm} \\ \hline
        WhiteFox~\cite{WhiteFox} & \makecell[l]{LM analysis,\\ fuzzing} & ~\redx & ~\redx & \makecell[l]{Memory, API misuse,\\ concurrency, logic,\\ security  vulnerabilities} & ~\redx & ~\greencheck$^{*}$ & ~\greencheck & ~\greencheck & ~\redx & \makecell[l]{TF XLA~\cite{tfxla},\\PyTorch \\Inductor~\cite{pytorch},\\
        LLVM~\cite{LLVM}}\\ \hline
        TorchProbe~\cite{Su_2023} & \makecell[l]{Feature-\\based fuzzing} & ~\redx & ~\redx & \makecell[l]{Memory, API misuse,\\ logic} & ~\redx & ~\greencheck$^{*}$ & ~\greencheck & ~\redx & ~\redx & \makecell[l]{PyTorch~\cite{pytorch},\\ Triton~\cite{TilletTriton}} \\ \hline
        \makecell[l]{MT-\\DLComp~\cite{Xiao_2022}} & \makecell[l]{Metamorphic\\ testing} & ~\redx$^{\dagger}$ & ~\redx & \makecell[l]{Memory, dead code,\\ layout optimization,\\ numeric accuracy} & \makecell[l]{Conv2D\\ layout\\ change} & ~\greencheck$^{*}$ & ~\greencheck & ~\redx & ~\redx & \makecell[l]{TVM~\cite{tvm},\\ Glow~\cite{Glow},\\ NNFusion~\cite{NNFusion}, \\TF XLA~\cite{tfxla}} \\ \hline
        FetaFix~\cite{FetaFix} & \makecell[l]{Fault localiza-\\tion,\\ differential\\ testing} & ~\redx$^{\dagger}$ & ~\greencheck & \makecell[l]{Preprocessing, input \\ dimensions, parameters,\\ hyperparameters, tensor\\ shape, graph structure} & ~\redx & ~\greencheck &~\redx & ~\redx & ~\redx & \makecell[l]{Converters across\\ PyTorch~\cite{pytorch},\\ Keras~\cite{chollet2015keras}, \\TF~\cite{tensorflow},\ TFLite~\cite{tensorflow}} \\ \hline
    \end{tabular}
    \label{t:related-work}
\end{table*}

In terms of the Technique column shown in Table~\ref{t:related-work},
most existing works primarily rely on compiler fuzzing, often enhanced with techniques such as differential testing (e.g., DeepDiffer~\cite{DeepDiffer}) or LLM analysis (e.g., WhiteFox~\cite{WhiteFox}). Consequently, these systems focus on triggering compiler bugs rather than pinpointing their origins and root causes, typically relying on generic compiler error messages to perform manual fault localization. Among these works, only FetaFix~\cite{FetaFix} provides a mechanism for fault localization, but it targets accuracy-related faults in model conversion between DL frameworks and does not support the model optimization process. In contrast, \tool\ systematically localizes faults by applying an iterative fault localization methodology for each pass in the ONNX Optimizer.

In terms of bug types detected in Table~\ref{t:related-work},
existing tools primarily focus on detecting crash bugs. For label accuracy-related bugs, they typically rely on random input generation (e.g., NNSmith~\cite{NNSmith}) to expose inconsistencies in model inference. In contrast, for label inconsistency bugs that \tool\ detects, we assess the prevalence and impact of these bugs on popular datasets(e.g., ILSVRC~\cite{ILSVRC17}, IMDB~\cite{imdbsentiment}) and model architectures.
Additionally, only a small number of tools consider miscompilations (for instance, models that contain problematic input nodes or misconfigured parameters), with just two prior works capable of detecting faults of this kind~\cite{TVMFuzz, WhiteFox}. 
The detection of bugs affecting execution times of the compiled models has received limited attention in the literature, with only two TVM-focused systems~\cite{TZER, DeepDiffer} using empirical measurements to evaluate performance-related faults. \tool\ currently identifies issues that could potentially lead to degraded model execution times by flagging redundant operator warnings (e.g., unused initializer entries).

Regarding support for ONNX in Table~\ref{t:related-work}, \tool\ is the only tool capable of localizing faults in the ONNX Optimizer. Other methodologies either treat ONNX merely as an intermediate representation or focus on bug detection in ONNX Runtime~\cite{WhiteFox}, a cross-platform inference engine, unlike model optimization targeted by the ONNX Optimizer.

Existing work on bug detection within graph optimization passes (shown as \textit{Gr. Opt. Passes} column in Table~\ref{t:related-work}) is  limited, with prior approaches focusing on common operations such as constant folding, operator fusion, and convolution optimizations. In contrast, \tool\ addresses the full spectrum of features in the ONNX Optimizer, covering a set of 47 passes that perform a wide variety of graph optimizations such as \texttt{Fuse}, \texttt{Eliminate}, and \texttt{Rewrite} operations in addition to the more common operations.

\vspace{-15pt}
\section{Experiment Setup}
\label{s:experiment-setup}
\subsection{The Experiment Process}
We evaluated all ONNX Optimizer passes (Section~\ref{sub:onnx-optimizer}). For each model, we first applied the default setting, which runs all fusion and elimination passes, and compared the outputs of the original and optimized models as described in Section~\ref{sub:comparison}. When discrepancies arose, we re-optimized using individual passes, producing 47 single-pass models to identify the pass causing the issue. \tool\ also supports new or custom passes, and our methodology generalizes to other optimization frameworks (e.g., Apache TVM) that allow both bundled and individual pass execution.

\subsection{DNN Models}
For our experiment set, we sourced $130$ model instances from the ONNX Model Hub~\cite{onnxmodelhub}.\footnote{\label{resultsfootnote}. The complete set of models used in our experiments, along with the full set of our results, is available in the \texttt{results} folder of the \tool\ repository at  \emph{\url{https://github.com/luludak/DiTOX/blob/master/results/README.md}}.} In particular, we selected $93$ models utilized for classification tasks (e.g., MobileNetV2~\cite{mobilenet}, 
EfficientNet-Lite4~\cite{efficientnet}
DenseNet121~\cite{densenet}, as well as InceptionV2~\cite{inception}), $30$ models used for object detection and semantic segmentation tasks (e.g., RetinaNet~\cite{retinanet}, Mask/Faster R-CNN R-50-FPN~\cite{mask-rcnn, faster-rcnn}, and SSD~\cite{ssd}), and $7$ transformer models~\cite{attention} utilized for text comprehension and generation tasks, such as question answering (Q\&A), (BERT-SQuAD~\cite{BERT}), sentiment analysis (RoBERTa~\cite{liu2020roberta}), text summarization (T5\\~\cite{T5Paper}) and generic-purpose text generation (GPT-2~\cite{gpt2}), with $4$ models focusing on Q\&A, while one model is utilized for each of the other text comprehension and generation tasks. We utilized models of a variety of opsets based on hub availability, ranging from $7$ (the lowest compatible with the latest ONNX Runtime version by the time of this publication - v$1.21.1$) up to $12$ (the highest provided from ONNX Model Hub via their API). Our methodology, however, is applicable to any ONNX model type and is not constrained by opset versions or tied to a specific model repository.

\subsection{Computational Resources}
We ran our models on mid-range systems with \texttt{Intel i5} and \texttt{i7} CPUs, reflecting the typical hardware used by developers during model development and fine-tuning.
Our method is generalizable to larger models, limited only by ONNX model size constraints, and can be extended to high-end systems with a trivial extension for hardware acceleration (using the ONNX Runtime API~\cite{onnxruntime}, or an AI compiler such as Apache TVM~\cite{tvm}).
\vspace{-10pt}
\subsection{Datasets}
We selected datasets appropriate to each model task. For classification and object detection, we used the \texttt{ILSVRC2017}~\cite{ILSVRC17} validation dataset, containing $5500$ images, commonly used for benchmarking image recognition models (e.g., ResNet~\cite{resnet} and BERT~\cite{BERT}). For text comprehension and generation, we used \texttt{SQuAD-v2}~\cite{rajpurkar2018knowdontknowunanswerable}, which includes $11873$ questions and contexts, to evaluate Q\&A tasks and general-purpose transformer behavior. Additionally, for summarization and sentiment analysis, we used the IMDB dataset~\cite{moviedataset, imdbsentiment}, containing $25000$ entries.

To accurately detect differences across model types, we employed comparison methods appropriate for each task. Specifically, we used the following approaches: \\

1. For image recognition models performing classification tasks, we utilized Kendall Rank Correlation Coefficient~\cite{kendall} (KT), measuring differences for top-K prediction labels of each dataset input image, for $K=1, 5$, and $10$.
If the KT value was found to be $< 1$, we treated this input as different. \\
\noindent 2. For image recognition models performing object detection and semantic segmentation, we calculated precision, recall, and F1 score for each dataset input, and extracted their average, combined with different thresholds of Intersection-over-Union (IoU) ($IoU=0.5, 0.75$, and $0.9$), as well as IoU, mean Average Precision (mAP), along with average precision and recall. We considered the original model output as the base, and the optimizer model the one under test for our metric  calculations, and the higher the values observed, the better the optimized model aligns with the original. \\
\noindent 3. For text generation models performing question answering and summarization, we utilized BLEU score~\cite{BLEU} across output texts for each input, monitoring instances where $BLEU\_Score < 1$ for each input. \\
\noindent 4. For sentiment analysis models, we compared each output value consisting of ${0, 1}$ across the original and the optimized versions, registering any differences observed. \\
\noindent 5. To distinguish actual faults from known optimizer limitations, we relied on ONNX Optimizer documentation and accounted for known limitations (e.g., split\_init and split\_predict are known to not fully comply with the ONNX standard format). Aside from issues from these two passes, all other detected behaviors were unexpected and thus classified as faults.

\section{Results}
\label{s:results}

Following the execution of the full set of experiments\footnotemark[1], \tool\ successfully identified $15$ bugs, reported across $11$ separate GitHub issues~\cite{githubissues}\footnote{An overview of all detected issues using \tool\ can be found at \emph{\url{https://github.com/luludak/DiTOX/blob/master/results/detected_issues.txt}}.}. A complete list of these issues, along with the associated models, is shown in Table~\ref{t:issues}. The developers have acknowledged the issues we reported, and we have provided them with full reproducibility details for each reported bug. Of the $15$ bugs identified, $14$ were entirely new (not previously reported on the ONNX Optimizer issue tracker), while one was a previously reported issue that lacked sufficient detail, including model information and reproducibility steps. We provided comprehensive information and full reproducibility details for this issue.

Within the ONNX optimizer, the bugs identified were primarily associated with $9$ of the $47$ passes. However, not all passes are equally important. By default, the optimizer prioritizes passes related to \texttt{fuse} and \texttt{eliminate} operations. As a result, $8$ of the $11$ reported GitHub issues are considered more critical since they belong to the core (default) strategy of the optimizer. Among these $8$ issues, $6$ are linked to specific passes (e.g., \texttt{fuse\_bn\_into\_conv}, which fuses batch normalization nodes with convolutions), while the remaining 2 issues were more general and related to the overall usage of the optimizer. We analyze and present the impact of these faults both on the optimization process and on the correctness of the optimized models in the following Sections.

\subsection{Optimizer Crashes}

Out of the $15$ detected bugs, $12$ bugs were related to model crashes, $2$ bugs resulted in output differences between the original and the optimized model, and $1$ bug introduced redundant initializers in the model that produced runtime warnings. In particular, we observed crashes were caused by $8$ out of the $47$ passes ($17$\%), and occurred for specific models, with object detection models being more prone to errors from these $8$ passes. In particular, \texttt{R-CNN} models, such as \texttt{Faster R-CNN R-50-FPN}~\cite{faster-rcnn}, \texttt{RetinaNet}~\cite{retinanet}, \texttt{VGG-16}, \texttt{VGG-19}~\cite{vgg}, as well as \texttt{YOLOv3}~\cite{Redmon2018YOLOv3AI}, presented crashes by using the optimizer, in a range of opset instances and in passes related to the modification of gather nodes, and the elimination of idempotent operations. In total, we observed $31$ model instances where either the optimizer crashed upon processing them, or the optimized model was problematic, leading to a runtime crash. Some secondary passes are known to be unstable (e.g., \texttt{split\_init} and \texttt{split\_predict} describe in their documentation that they not strictly comply with the ONNX format), and therefore we do not focus on them as sources of faults in the optimization process. Overall, $12$ out of the $31$ model instances ($9.2$\%) that were associated with a crash occurred while using passes related to \texttt{fuse} and \texttt{eliminate} operations, which are part of the default bundle of optimizations in the ONNX Optimizer and are therefore of particular significance.

As shown in Table~\ref{t:issues}, crash faults can be attributed to three main causes: (1) model input issues (cases \#$4$ and \#$6$), (2) incorrect use of initializers (cases \#$12$ and \#$13$), and (3) issues associated with node properties such as duplicate names, obsolete fields, and incorrect dimensions (all other cases). Overall, object detection models were more prone to optimizer crashes compared to classification models. Due to their increased structural complexity and larger size, they are more likely to exhibit errors related to node properties and initializers, which account for the vast majority of crash faults. In contrast, only one issue was observed in the most complex model category — text generation models, that was related to model input in GPT-2. However, upon further inspection, we found that the transformation passes of the ONNX Optimizer had limited impact on text generation models — the majority of optimization passes left the models unchanged, and those that did make modifications affected a small proportion of nodes with negligible effect on outputs.

\setlength{\tabcolsep}{1.5pt}
\begin{table*}[!htp]
    \centering
    \small
    \caption{
    Overview of the issues identified using \tool, detailing the responsible optimization pass, the detected error, its type (Crash, Accuracy (Acc.), or Warning (Warn.)), and the models in which the error was observed, along with their opsets (op.). The (Run) label in crashes indicates that the model was successfully optimized but failed during execution.}
    \label{t:issues}
    \begin{tabular}{|l|l|l|l|l|}
    \hline
        \textbf{\#} & \textbf{Pass} & \textbf{Error} & \textbf{Type} & \textbf{Models} \\ \hline
        1 & lift\_lexical\_references & \makecell[l]{Removes used value references.} & Crash & \makecell[l]{EfficientNet-Lite4 (op. 11), SSD (op. 12)} \\ \hline
        2 & split\_init & \makecell[l]{Out of range List index} & Crash & YOLOv3 (op. 10, 12) \\ \hline
        3 & \makecell[l]{split\_init\\
        split\_predict} & \makecell[l]{Missing required field} & Crash & \makecell[l]{
        DenseNet121 (op. 7-9), 
        YOLOv2 (op. 10),\\ 
        SSD (op. 12),
        Tiny-YOLOv3 (op. 11),  
        YOLOv3 (op. 10, 12)} \\ \hline
        4 & split\_predict & \makecell[l]{Missing required inputs} & Crash &
        \makecell[l]{
        SSD (op. 12),  DenseNet121 (op. 7), 
        InceptionV2 (op. 6-9)} \\ \hline
        5 & split\_predict & Bad spec for node. & Crash & \makecell[l]{
        YOLOv3 (op. 10),
        DenseNet121 (op. 7),\\
        InceptionV2 (op. 6-9) } \\ \hline
        6 & split\_predict & \makecell[l]{Mismatch in \#input elements} & Crash & \makecell[l]{FCN ResNet-50-int8 (op. 12)} \\ \hline
        7 & eliminate\_shape\_gather & \makecell[l]{Indices value $\neq$ node dims size.} & Crash & \makecell[l]{Faster R-CNN: R-50-FPN (op. 10)} \\ \hline
        8 & All Passes & \makecell[l]{Implicit ONNX format version update.} & Acc. & \makecell[l]{Models of cases $1$-$11$ in  Table~\ref{t:classification}}\\ \hline
        9 & rewrite\_input\_dtype & \makecell[l]{Multiple usage of output name} & Crash & GPT-2 (op. 10) \\ \hline
        10 & \makecell[l]{eliminate\_consecutive\\\_idempotent\_ops} & \makecell[l]{Incompatible shape error.} & \makecell[l]{Crash\\(Run)} & \makecell[l]{
        Mask R-CNN/Faster R-CNN: R-50-FPN-int8 (12),\\
        R-50-FPN-qdq (op. 12),
        R-50-FPN-fp32 (op. 12)} \\ \hline
        11 & \makecell[l]{eliminate\_duplicate\_initializer} & \makecell[l]{Erase output assertion failed.} & Crash & \makecell[l]{Faster R-CNN: R-50-FPN-qdq (12)} \\ \hline
        12 & fuse\_bn\_into\_conv & \makecell[l]{Duplicate initializer\\ introduced.} & \makecell[l]{Crash\\(Run)} & RetinaNet (op. 9) \\ \hline
        13 & \makecell[l]{eliminate\_shape\_gather\\eliminate\_nop\_reshape} & \makecell[l]{Unused initializers} & Warn. & \makecell[l]{Faster R-CNN: R-50-FPN-int8 (op. 12)} \\ \hline
        14 & All Passes & Invalid input scale. & Crash & \makecell[l]{VGG-16-BN (op. 7), VGG-19-BN (op. 7), \\ResNet101\_DUC\_HDC (op. 7)} \\ \hline
        15 & fuse\_bn\_into\_conv & Bad graph structure. & Acc. & \makecell[l] { Efficient-Lite4 (op. 11),  SSD (op. 12) } 
        \\ \hline
    \end{tabular}
\end{table*}

\subsection{Output Label Differences in Optimized Classification \& Object Detection Models}
Out of the $130$ optimized models tested, we observed unexpected output differences between the original and optimized models. These differences manifested as (1) label differences for classification models (e.g., ResNet), (2) label and/or bounding box differences for object detection models (e.g., YOLOv3), and (3) textual output differences in token sequences or predicted spans for text generation tasks. Since the ONNX Optimizer transformations are graph-based and do not alter numeric precision, output differences are not expected. Among the $93$ classification models tested, $28$ ($30.1$\%) exhibited discrepancies in output labels. Similarly, $5$ out of $30$ ($16.6$\%) object detection models presented output differences.
The text generation models, on the other hand, were robust with only one instance (GPT-2) exhibiting negligible differences ($< 0.001\%$) while all others showing identical results across the original and the optimized model versions.
The differences impacted the top-5 and top-10 predictions, while top-1 predictions remained unchanged. Although top-1 accuracy is often emphasized, variations in lower-ranked predictions can also reveal important issues. Our results show that focusing exclusively on top-1 accuracy may give developers a false sense that an optimization is successful, while overlooking substantial changes in top-K predictions (up to 2\% in top-5 and 13\% in top-10). Using \tool, we detected these inconsistencies and reported them to the ONNX Optimizer developers.

\begin{table}[!htp]
    \centering
    \caption{
    Output label accuracy differences across classification (CL) and object detection (OD) models, showing type, opset version (\textit{Ver.}), and accuracy gaps for top-1 (\textit{T-1}), top-5 (\textit{T-5}), and top-10 (\textit{T-10}) predictions between original and ONNX-optimized models, along with the related issue (Is.): Format Version (FV) or Malformed Graph (MG).}
    \small
    \begin{tabular}{|l|l|l|l|l|l|l|l|}
    \hline
        \textbf{\#} & 
        \textbf{Model} & \textbf{Type} & \textbf{Ver.} & \textbf{T-1} & \textbf{T-5} & \textbf{T-10} & \textbf{Is.} \\ \hline
        1 & AlexNet~\cite{alexnet} & CL & 7-9 & 0\% & 0\% & \cellcolor{orange!20}0.02\% & FV \\ \hline
        2 & DenseNet121~\cite{densenet} & CL & 7-9 & 0\% & 0\% & \cellcolor{orange!20}0.02\% & FV \\ \hline
        3 & GoogleNet~\cite{googlenet} & CL & 7-9 & 0\% & 0\% & \cellcolor{orange!25}0.04\% & FV \\ \hline
        4 & InceptionV1~\cite{inception} & CL & 7-9 & 0\% & 0\% & \cellcolor{orange!50}0.25\% & FV \\ \hline
        5 & InceptionV2~\cite{inception} & CL & 7-9 & 0\% & \cellcolor{orange!20}0.02\% & \cellcolor{orange!25}0.05\% & FV \\ \hline
        6 & MobileNetV2~\cite{mobilenet, mobilenetv2} & CL & 7-9 & 0\% & 0\% & \cellcolor{orange!20}0.02\% & FV \\ \hline
        7 & ResNet18~\cite{resnet} & CL & 7 & 0\% & \cellcolor{orange!20}0.02\% & \cellcolor{orange!20}0.02\% & FV \\ \hline
        8 & ResNet34V2~\cite{resnetv2} & CL & 7 & 0\% & \cellcolor{orange!20}0.02\% & \cellcolor{orange!20}0.02\% & FV \\ \hline
        9 & ResNet50-caffe2~\cite{resnet} & CL & 7-9 & 0\% & \cellcolor{orange!70}2\% & \cellcolor{orange!100}13\% & FV \\ \hline
        10 & ShuffleNetV1~\cite{shufflenet} & CL & 7-9 & 0\% & \cellcolor{orange!30}0.7\% & \cellcolor{orange!60}1\% & FV \\ \hline
        11 & SqueezeNet 1.0~\cite{squeezenet} & CL & 7-9 & 0\% & 0\% & \cellcolor{orange!20}0.02\% & FV \\ \hline
        12 & EfficientNet-Lite4~\cite{efficientnet} & CL & 11 & 0\% & \cellcolor{orange!35}0.04\% & \cellcolor{orange!60}1\% & MG \\ \hline
        13 & SSD~\cite{ssd} & OD & 12 & 0\% & 0\% & \cellcolor{orange!35} 0.04\% & MG \\ \hline
    \end{tabular}
    \label{t:classification}
\end{table}

\begin{figure}[!ht]
 \centering
 \includegraphics[width=\linewidth]{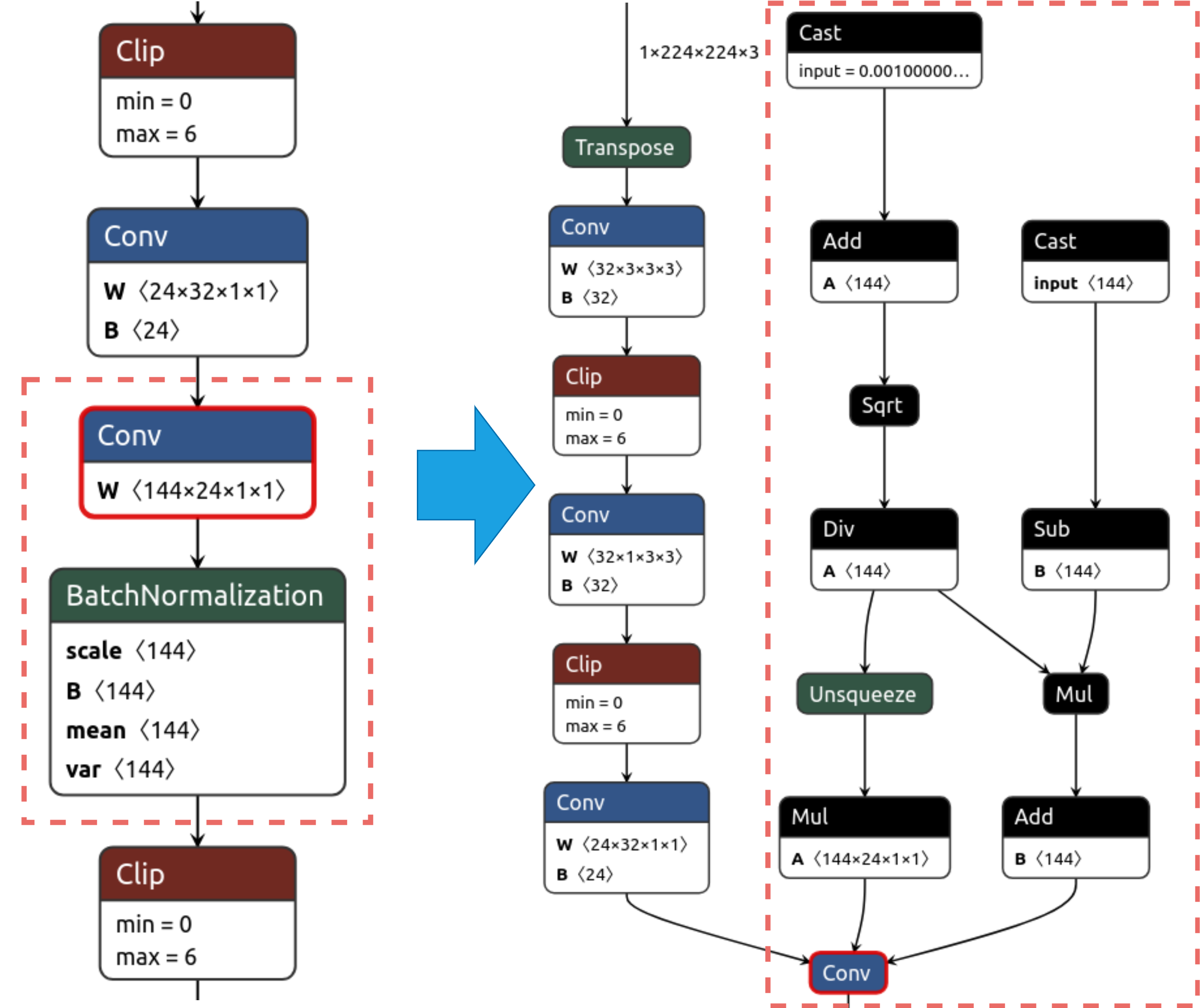}
 \caption{Incorrect model graph transformation caused by the \texttt{fuse\_bn\_to\_conv} pass in EfficientNet-Lite-4. The nodes affected are highlighted. It should produce a \texttt{Conv} node with updated weights and biases, but the pass generates multiple nodes that represent the \texttt{BatchNormalization} formula.}
 \label{fig:bnconvbug}
 \vspace{-10pt}
\end{figure}

Overall, we detected two primary issues related to output label differences between the original and optimized models:

\begin{description}[leftmargin=0pt]
    \item[Implicit ONNX format version upgrade:] The ONNX Optimizer silently upgrades models from ONNX format version \texttt{v3} to \texttt{v4}, without any warning or indication. This behavior impacts models using opset versions 7, 8, and 9— that are older opsets but still supported in the current ONNX Runtime (\texttt{v1.21.1}). Despite being outdated, these opset versions are still prevalent, representing approximately $41$\% of the models in the ONNX Model Hub (presented in Table~\ref{t:classification}, with the issue labeled as ``FV", indicating a problem with the format version.). While not all models within this opset range are affected, we observed a significant number that exhibited differences between their original and optimized versions. We have reported this issue to the ONNX Optimizer developers, who have acknowledged it and are investigating further.
    \vspace{5pt}
    \item[Problematic fusion pass:] The pass \texttt{fuse\_bn\_into\_conv\\}  fuses Batch Normalization and Convolution operators within the model. In particular, the pass attempts to fuse \texttt{BatchNor-\\malization} nodes (which do not update their mean or variance during inference)  with their preceding \texttt{Conv} nodes by adjusting the weights and biases of the \texttt{Conv} node according to the properties of the \texttt{BatchNormalization} node. We observed that this pass was problematic in a number of cases, resulting in both a crashing bug and an issue where redundant nodes are introduced into the model. In the latter scenario, instead of properly merging the nodes resulting in a \texttt{Conv} node with adjusted weights and biases, the pass generates multiple nodes that represent the Batch Normalization formula. This behavior is illustrated on the right side of Figure~\ref{fig:bnconvbug}. This formula representation is problematic as several models produced different output labels after optimization using this pass compared to their original versions (presented in Table~\ref{t:classification}, with the issue labeled as ``MG'', indicating a malformed graph). Although another user previously reported a similar issue, no details on the reproduction steps or affected models were provided. We submitted a reproducible issue to the ONNX Optimizer developers, including all relevant information and behavior details.
\end{description}

Table~\ref{t:classification} shows label differences in top-K outputs for classification and object detection models, highlighting the causes. 

\setlength{\tabcolsep}{1.5pt}
\begin{table}[!ht]
    
    \caption{Object detection metric differences between the original and optimized versions, using the default ONNX Optimizer setting. For each model, we present the average F1 score (\textit{F1}), precision (\textit{Prec.}), and recall (\textit{Rec.}), as well as mAP and IoU.}
    \vspace{-10pt}
    \fontsize{10}{10}\selectfont
    \label{t:ODScores}
        \begin{tabular}{|l|l|l|l|l|l|l|}
        \hline
            \textbf{Model} & \textbf{Op.} & \textbf{F1} & \textbf{Prec.} & \textbf{Rec.} & \textbf{mAP} & \textbf{IoU} \\ \hline
            SSD~\cite{ssd} & 12 & 99.99\% & 99.99\% & 99.99\% & 99.99\% & 100\% \\ \hline
            \makecell[l]{Tiny\\ YOLOv3~\cite{Redmon2018YOLOv3AI}} & 11 & \cellcolor{orange!35}99.18\% & \cellcolor{orange!35}99.18\% & \cellcolor{orange!35}99.18\% & 99.99\% &  \cellcolor{orange!70}89.35\% \\ \hline
            YOLOv2~\cite{yolov2} & 9 & \cellcolor{orange!20}99.95\% & \cellcolor{orange!20}99.95\% & \cellcolor{orange!20}99.95\% & \cellcolor{orange!20}99.94\% & \cellcolor{orange!70}88.49\% \\ \hline
            YOLOv3~\cite{Redmon2018YOLOv3AI} & 10 & \cellcolor{orange!20}99.45\% & \cellcolor{orange!20}99.45\% & \cellcolor{orange!20}99.45\% & 99.99\% & \cellcolor{orange!80}84.81\% \\ \hline
            YOLOv3~\cite{Redmon2018YOLOv3AI} & 12 & \cellcolor{orange!20}99.45\% & \cellcolor{orange!20}99.45\% & \cellcolor{orange!20}99.45\% & 99.99\% & \cellcolor{orange!80}84.81\% \\ \hline
        \end{tabular}
        \vspace{-10pt}
\end{table}

\hspace*{-10pt}\textbf{Bounding Box Differences in Object Detection Models}
We observed some differences in the bounding box outputs between the original and optimized versions of object detection models. The results are presented in detail in Table~\ref{t:ODScores} with the following main findings: 1. Both \texttt{YOLOv3} models (opset $10$ and $12$) achieve high average IoU, F1 score, precision, and recall, all above $99.45$\%. However, the optimized model occasionally misplaces bounding boxes—resulting in IoU drops of up to $15.19$\%, these differences do not lead to misclassifications, as none were observed in the top ranked label predictions. A similar trend is also seen in \texttt{YOLOv2} model where optimization causes a drop in IoU value but not in detection performance (F1, precision and recall). 
Although this behavior may appear minor, it can lead to undesirable consequences, particularly in safety-critical systems where even small deviations matter. While our experiments did not observe bounding box misclassifications, such outcomes cannot be ruled out for other models or different inputs.

\subsection{Optimizer Warnings}

We observed two issues related to the addition of redundant elements (unused initializer entries), introduced in the optimized models, associated with the \texttt{eliminate\_shape\_gather} and \texttt{eliminate\_nop\_reshape} passes. When executed with ONNX Runtime, these entries were ignored, triggering a warning in the format: \texttt{Removing initializer `INITIALIZER\_ID'. It is not used by any node and should be removed from the model.} Upon manual inspection, we found no significant differences in model size. However, this issue warrants further investigation for potential side effects \texttt{towards model performance}. We have reported it to the optimizer developers for additional review.

\section{Tool Usability}
\tool~ can test whether the ONNX Optimizer generates optimized models without compromising output label correctness.
Currently, \tool\ supports differential testing for classification, object detection, and text generation models. However, \tool\ can also be easily extended to support new model types, as long as (1) the models can be compiled into the ONNX format for optimization and (2) a comparator is provided to compare the results between the base and optimized models for the new model type.
In addition,  \tool\ can fetch models and datasets automatically, making it adaptable for use in a Continuous Integration pipeline. This allows \tool\ to test the ONNX Optimizer whenever new functionality (e.g., new passes) is added. By default, \tool\ loads models from the official ONNX Model Hub~\cite{onnxmodelhub}, which was used in our experiments. However, by configuring \tool\ to use a custom ONNX Hub that follows the ONNX API specification for model repositories~\cite{onnxhubapi}, it can fetch and use models without further modification. The same flexibility applies to datasets. \tool\ can fetch datasets using packages like the \texttt{datasets}~\cite{datasets} package or from local storage.

\section{Tool Generalizability}
Although \tool\ applies differential testing to the ONNX Optimizer, the approach is generic and can be adapted to other AI optimizers or compilers that provide a runtime for model execution and expose optimization passes both collectively and individually through an API. Many compiler ecosystems already satisfy these requirements. For example, MLIR~\cite{MLIR} offers passes such as graph rewrites, Apache TVM~\cite{tvm} provides a Sequential API for composing optimizations like constant folding and subexpression elimination, and OpenVINO~\cite{openvino} supports a full set of individual passes. While we focus on ONNX due to its wide adoption, the methodology of \tool is model-format independent and can be applied to other frameworks using their supported model types and runtimes. Finally, \tool\ includes comparators for classifiers, object detectors, and transformers; for other model families, users can implement custom comparators by following the already provided comparator API.

\section{Threats To Validity}
\noindent We list the threats to validity in our evaluation below:  

\textbf{Model Selection:}
Our evaluation covered 130 models from the official ONNX Model Hub, spanning classification, object detection, as well as text comprehension and generation. Models were chosen for their maturity, accessibility, and broad adoption. aimed to create a representative and diverse experimental set, although the selection is not exhaustive. We also excluded models with opset versions below 7 due to incompatibility with current ONNX and ONNX Runtime versions, limiting our results to newer models. To keep the study tractable, diffusion and reinforcement learning models were omitted; we plan to include these in future work.

\textbf{Methodology Generalizability:}
We focus ONNX Optimizer because it is a core component of the widely used ONNX ecosystem (19.5K stars)~\cite{onnxsite} and supports tasks beyond model optimization, such as graph simplification (ONNX Simplifier, 4.1K stars~\cite{onnxsimplifier}). Its popularity and broad applicability make it a strong candidate for study and a representative of the state-of-the-art technologies. While testing all DL compilers was infeasible, we demonstrated the value of our methodology and noted it can extend to other optimizers that offer per-pass APIs and runtime support for differential testing. For example, DiTOX could support OpenVINO~\cite{openvino} and Apache TVM~\cite{tvm} by adapting their graph optimization passes, considering both compilers also provide runtime APIs, similarly to ONNX.

\textbf{Optimization Pass Ordering:}

The order of optimization passes can significantly affect model performance and behavior~\cite{OptimizationOrder}, a well-known challenge in compiler research. While we do not study pass ordering in depth, \tool\ adopts two evaluation modes: (1) the default ONNX Optimizer configuration, including common passes such as \texttt{fuse} and \texttt{eliminate}, and (2) a configuration applying all 47 available passes atomically. Although this does not capture issues from specific pass combinations or orderings, it effectively uncovers correctness problems in practice. Nevertheless, the omission of a systematic study on optimization ordering may limit the completeness of our evaluation.

\vspace{-10pt}
\section{Discussion}
\noindent We summarize our findings across different model types below. 

\noindent \textbf{(1) Optimizations can affect label accuracy in classification models.} 
Label discrepancies in classification models arose primarily from two sources. First, models using legacy ONNX opsets were implicitly and incorrectly upgraded by the optimizer, disproportionately affecting classification models. Second, the \texttt{fuse\_bn\_into\_conv} pass occasionally generated problematic graph nodes, altering model behavior in convolution-heavy architectures. In contrast, text models were largely unaffected, as the optimizer applies few passes to transformer-based architectures where convolutional patterns are rare.

\noindent \textbf{(2) Object detection models are more prone to optimization crashes.} 
Most optimizer crashes occurred in object detection and semantic segmentation models. Their larger size and structural complexity trigger more aggressive transformations, increasing exposure to corner cases in optimization passes.

\noindent \textbf{(3) Optimization can degrade bounding box accuracy.}
While overall detection accuracy remained stable, optimized models occasionally produced misaligned bounding boxes. This resulted in IoU drops of up to $15.19$\%, despite negligible changes in top-10 accuracy (e.g., $0.04$\% for SSD).

\textbf{(4) Text models are largely robust to optimization.} 
Among the eight text comprehension and generation models, we observed only one optimizer crash (\texttt{GPT-2}) and minor BLEU-score differences affecting eight samples in top-3 token predictions.

Overall, ONNX optimizations can introduce correctness issues, with susceptibility varying by model type. \tool\ effectively isolates the optimization passes responsible for these discrepancies across model classes.

\section{Future Work}
We identify three potential extensions of this work as future research directions:

\textbf{(1) Execution Time Measurements:}
This study primarily focused on optimizer correctness, specifically on detecting crashes and accuracy faults. However, in the broader context of model optimization, it is also important to assess how an optimizer impacts overall model performance. In particular, this includes execution time for both CPU and GPU deployments, as well as other factors such as model size. \tool\ can be easily extended to support hardware-accelerated model execution, either by leveraging the ONNX Runtime API~\cite{onnxruntime} or by integrating a third-party AI compiler that generates target-specific device code, such as Apache TVM~\cite{tvm}. We leave this direction for future work, as it could provide valuable additional insights.

\textbf{(2) Selection of Different Model Types:}
This work focused on classification, object detection, and text generation models. However, \tool\ is applicable to a broader range of model types, including speech recognition, super-resolution, and body recognition. We leave the exploration of these additional model categories as a direction for future work.

\textbf{(3) Examination of the Optimization Pass Order:}\\
\tool\ supports optimization using a user-defined subset and ordering of passes through configuration. Although such experiments could potentially reveal additional faults, we exclude them for tractability, as exhaustively generating, executing, and evaluating all pass permutations is computationally infeasible. We leave this direction for future work, potentially in combination with techniques for reducing the permutation search space.

\vspace{-10pt}
\section{Conclusion}
We propose \tool, a tool for the differential testing of the ONNX Optimizer. Using \tool, we evaluated $130$ models across classification, object detection, and text generation tasks. We detected $9.2$\% of model instances that crashed when using the primary optimizer strategies, while $30$\% of classification models and $16.6$\% of object detection models showed differences between the source and optimized versions, while text generation models remained robust. Additionally, \tool\ identified 15 issues, 14 of them new, related to crashes and accuracy degradation across 9 optimizer passes, which we reported to the ONNX Optimizer developers.


\balance
\bibliographystyle{ACM-Reference-Format}
\bibliography{00-main}

\end{document}